\documentclass[runningheads]{llncs}
\usepackage[T1]{fontenc}
\usepackage{graphicx}
\usepackage{amssymb}
\usepackage{multirow}
\usepackage{booktabs}
\usepackage{subfigure}
\usepackage{subfloat}
\usepackage{tabularx} 
\usepackage{caption}
\usepackage[misc]{ifsym}
\begin{document}

\title{FRCNet: Frequency and Region Consistency for Semi-supervised Medical Image Segmentation}

\author{Along He\inst{1} \and Tao Li \inst{1,2}$^{(\textrm{\Letter})}$ \and Yanlin Wu \inst{1} \and  Ke Zou \inst{3} \and Huazhu Fu \inst{4}}
\authorrunning{F. Author et al.}

\institute{Tianjin
	Key Laboratory of Network and Data Security Technology, College of Computer Science, Nankai University, Tianjin, China \\ 
		 \email{litao@nankai.edu.cn}\\ \and
	Haihe Lab of ITAI,\\
    \and  National Key Laboratory of Fundamental Science on Synthetic Vision,
    Sichuan University, Chengdu, Sichuan, China\\  \and 
Institute of High Performance Computing, Agency for Science, Technology and
Research, Singapore, Singapore}

\maketitle              

\begin{abstract}
	Limited labeled data hinder the application of deep learning in medical domain. In clinical practice, there are sufficient unlabeled data that are not effectively used, and semi-supervised learning (SSL) is a promising way for leveraging these unlabeled data.  However, existing SSL methods ignore frequency domain and region-level information and it is important for lesion regions located at low frequencies and with significant scale changes. In this paper, we introduce two consistency regularization strategies for semi-supervised medical image segmentation, including frequency domain consistency (FDC) to assist the feature learning in frequency domain and multi-granularity region similarity consistency (MRSC) to perform multi-scale region-level  local context information feature learning.  With the help of the proposed FDC and MRSC, we can leverage the powerful feature representation capability of them in an effective and efficient way. We perform comprehensive experiments on two datasets, and the results show that our method achieves large performance gains and exceeds other state-of-the-art methods.
\end{abstract}

\section{Introduction}

		Medical image segmentation aims to segment the anatomical structure, organs or lesions from the medical images for clinical analysis and diagnosis such as skin disease diagnosis \cite{Wu2021AutomatedSL} and fundus screening \cite{Fu2018DiscAwareEN}. It has been significantly developed with the rise of convolutional neural networks (CNNs) \cite{ronneberger2015u,zhou2018unet++,li2018h}. While existing approaches rely on high-quality labeled data, which is labor-intensive and time-consuming in the medical domain. It is not always feasible to collect sufficient labeled data in real world applications, especially for pixel-wise segmentation tasks. 
		
		Training with limited data may easily lead to over-fitting and catastrophic forgetting \cite{Kirkpatrick2016OvercomingCF}, causing the performance drop. In clinical practice, there are a large number of unlabeled data available, and the dependence on labeled data can be alleviated if these data can be utilized effectively. Therefore, it is promising to develop semi-supervised learning (SSL) methods to reduce the burden of annotation in medical domain. Therefore, many attempts have been made at SSL \cite{tarvainen2017mean,sohn2020fixmatch}. They introduce data-level \cite{li2020transformation,wu2021semi} and model-level consistency \cite{tarvainen2017mean,ouali2020semi,luo2021efficient}, or generate pseudo labels using the model’s predictions \cite{chen2021semi}.  
		
		 Despite achieving good results, previous methods still face the following limitations. \textbf{Firstly}, current SSL methods mainly focus on RGB domain, ignoring the frequency domain. The intensity of most lesion regions in medical images is flat and they are low-frequency signals \cite{zhong2022detecting}. Segmenting lesions with only the RGB domain information is an challenging task without explicitly considering frequency information. Therefore, we aim to explore the rich information of frequency and RGB domains.  \textbf{Secondly}, most of previous methods focus on pixel-level consistency, and lack semantic consistency at region level. Considering multiple region-level consistency can effectively deal with lesions of different sizes and enhance the multi-scale features modeling ability.  
		
		In this paper, we come up with two consistency regularization strategies for SSL, one is frequency domain consistency (FDC), and the other is multi-granularity region similarity consistency (MRSC). For FDC, we apply the Discrete Cosine Transform (DCT)  \cite{ahmed1974discrete} to convert the RGB domain features into the frequency domain. To discover the cues of lesion information in frequency space, we design a frequency enhancement module (FEM) with Transformer \cite{vaswani2017attention} to encode the features in frequency domain, which can boost the capabilities of the model in processing frequency signals. For MRSC, we aim to provide multi-granularity regional context information, rather than only pixel-wise consistency. We adopt the Mean Teacher (MT) \cite{tarvainen2017mean} framework as base framework, and the proposed consistency regularization is incorporated to form the Frequency and Region Consistency (FRCNet) for SSL medical image segmentation. 
		
		The main contributions of our work: (1) We propose FDC and MRSC consistency regularization strategies, which can learn from sufficient unlabeled data to deal with insufficient annotation.  
		(2) Our method is plug-and-play and it can be easily combined with existing SSL methods. In the inference time, FDC and MRSC can be safely removed, reducing the inference complexity. 
		(3) Finally, our FRCNet outperforms the state-of-the-art (SOTA) methods on two datasets by a large margin.

\section{Method}
\subsection{Notations and Overview.}
	   For SSL, training set contains $M$ labeled and $N$ unlabeled data, where $N \gg M$. The labeled pairs and unlabeled images are denoted as $\mathcal{D}^l = \{(x_i,y_i)\}_{i=1}^M$ and  $\mathcal{D}^u = \{x_j\}_{j=1}^N$, respectively, where $i$ and $j$ are indexes of labeled and unlabeled data, respectively, and the training dataset is $\mathcal{D} = \mathcal{D}^l \cup \mathcal{D}^u$. 

	   \begin{figure}
	   	\centering
	   	\includegraphics[width=12cm]{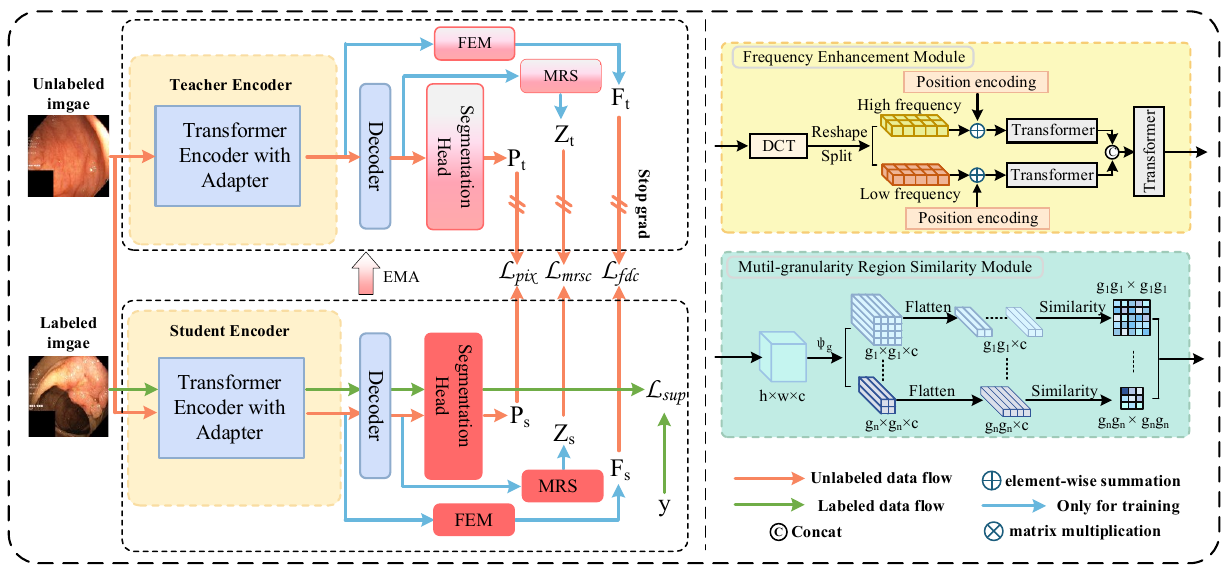}
	   	\caption{Overall framework of the proposed FRCNet, which is based on MT structure.  For labeled data, they  are used to train the student network directly with the labels, and unlabeled data is used to train the student network through the proposed consistency regularization.} \label{structure}
	   \end{figure}

		The framework of our method is illustrated in Fig \ref{structure}, mean teacher (MT) \cite{tarvainen2017mean} is employed as the baseline,  which consists of a teacher $F_t$ and a student model $F_s$. 
		For $F_s$, it contains a Transformer encoder with adapter \cite{chen2022adaptformer}, a decoder, a segmentation head, MRS and FEM modules. 
		MRS is to obtain multi-scale region-level features, and the FEM branch is used to transform the RGB features to frequency domain. Formally, the outputs of teacher and student models are denoted as: segmentation prediction $\mathcal{P}$, frequency domain features $\mathcal{F}$ and regional features $Z$. For labeled data, student model is trained with cross-entropy and Dice loss, they are formulated as:
		\begin{equation}
			\mathcal{L}_{sup} = \frac{1}{2|\mathcal{D}_l|} \sum_{(x_i,y_i)\in \mathcal{D}_l} \mathcal{L}_{ce}(y_i,\mathcal{P}^s_i)+\mathcal{L}_{dice}(y_i,\mathcal{P}^s_i),
			\label{sup_s}
		\end{equation}
		where $s$ denotes student model.
		
		For unlabeled data, student model can be optimized through the pixel-level loss, and the proposed frequency domain and region-level consistency, which is described in Fig \ref{structure}.  During training, we only fine-tune the parameters of segmentation head, adapters, FEM and MRS for parameter efficiency. Exponential moving average (EMA) is used to update teacher model  \cite{tarvainen2017mean}.
				
\subsection{Frequency Domain Consistency.}
		For RGB features from the last stage of Transformer encoder  $X \in \mathbb{R}^{h\times w\times d}$, where $h$, $w$ and $d$ denote the height,  width and channel number, respectively. First, we transform the RGB  features $X$ using DCT  \cite{ahmed1974discrete}  and obtain frequency features. Then, we divide it into $N$ patches of size $ p\times p$, and $ N = \frac{h\cdot w}{p^2}$, and the frequency domain features are denoted as $\mathcal{F}_{raw} \in \mathbb{R}^{\frac{h}{p}\times \frac{w}{p}\times c}, c = p \cdot p\cdot d$, and thus the RGB features are transformed to the frequency domain, and $p$ is set to 2.
		
		\textbf{Frequency Enhancement Module.} After the transformation of DCT, we adopt Transformer \cite{vaswani2017attention} to enhance the raw frequency features, as shown on the right of Fig \ref{structure}. To model lesion features that belong to low-frequency, we decompose the raw frequency domain $\mathcal{F}_{raw} = \{\mathcal{F}_l,\mathcal{F}_h\}$, where $\mathcal{F}_l \in \mathbb{R}^{\frac{h}{p}\times \frac{w}{p}\times \frac{c}{2}}$ and $\mathcal{F}_h  \in \mathbb{R}^{\frac{h}{p}\times \frac{w}{p}\times \frac{c}{2}}$ denote the low-frequency component  and the high-frequency component, respectively. To boost the signals in the frequency domain, the low-frequency component $\mathcal{F}_l$ and high-frequency component $\mathcal{F}_h$ are fed into two independent Transformer blocks to encode frequency features, which can be denoted as 
		\begin{equation}
			\mathcal{F}_l' = \Phi_1(\mathcal{F}_l + E^{pos}_l),   \mathcal{F}_h' = \Phi_2(\mathcal{F}_h + E^{pos}_h),
		\end{equation}
		where $\Phi_1$ and $\Phi_2$ are the two Transformer blocks, each of them consists of a multi-head self-attention layer and multi-layer perceptron. $E^{pos}_l \in \mathbb{R}^{\frac{h}{p}\times \frac{w}{p}\times \frac{c}{2}}$ and $E^{pos}_h \in \mathbb{R}^{\frac{h}{p}\times \frac{w}{p}\times \frac{c}{2}}$ are the learnable position embeddings to preserve the frequency location information. Then, another Transformer block $\Phi_3$ is adopted to encode all the low-frequency and high-frequency components to better fuse the features in frequency space: $\mathcal{F} = \Phi_3([\mathcal{F}_l', \mathcal{F}_h' ])$,   
		$[. , .]$ denotes concatenation along the channel dimension, and $\mathcal{F}$ is the final fused high and low frequency features. 
		
		\textbf{Frequency Domain Consistency Loss.} For the same input images, after the feature encoding in frequency domain, the outputs of student and teacher models should be consistent. We can obtain the unsupervised consistency loss on unlabeled data. 
		For simplicity, we use Mean Square Error (MSE) loss as the consistency regularization:
		\begin{equation}
			\mathcal{L}_{fdc} =\frac{1}{|\mathcal{D}_u|} \sum_{x_j \in \mathcal{D}_u} \left \| \mathcal{F}^t_{j} -\mathcal{F}^s_{j} \right \|_2^2,
		\end{equation}
		where $\mathcal{L}_{fdc}$ is the frequency domain consistency loss, $\mathcal{F}^t_{j}$ and $\mathcal{F}^s_{j}$ are the frequency space features of teacher and student models, respectively.
		
\subsection{Multi-granularity Region Similarity Consistency.} 
 		  Although the SSL methods based on pixel-level consistency have achieved decent results, they ignore the local information in 2D images, and thus they cannot model the relationship between local regions well. Therefore, we propose multi-granularity region similarity consistency, which builds region-level contextual relationships with different scales, as shown on the right of Fig. \ref{structure}.  
 		  
 		  Specifically, in order to obtain the features at region level, we adopt the output from the last stage of decoder as the region features. The features from the teacher and student models are denoted as $Z^t$ and $Z^s \in \mathbb{R}^{h'\times w'\times c'}$. First, they are projected and flatten into small regional features $R$, 
 		  \begin{equation}
 		  	R = \Gamma (\Psi_g(Z)) \in \mathbb{R}^{gg \times c'}, 
 		  \end{equation}
 		  where $\Gamma$ is the flatten operation and $\Psi_g$ is the region projection with the granularity of $g$. The region $R$ is with the size of  $g \times g$, and each element in $R$ represents a local region in original image. It can be implemented with arbitrary downsampling methods, such as linear interpolation and pooling based methods, and we use linear interpolation in our method (See Table. \ref{table_region}). Then, we compute  region similarity  matrix $A = RR^T \in \mathbb{R}^{gg\times gg}$, where the elements indicates the similarity between regions, they can be used to model the similarity of region features. The region similarity consistency loss is to minimize the difference between the teacher and student model:
 		  \begin{equation}
 		  	\mathcal{L}_{mrsc} = \frac{1}{|\mathcal{D}_u|} \sum_{x_j \in \mathcal{D}_u} \sum_{g \in G}\left \| A_{j,g}^t - A_{j,g}^s \right \|_2^2,
 		  \end{equation}
 		  $G$ denotes the total granularities, $A_{j,g}^t$ and $  A_{j,g}^s$ are  region similarity scores from teacher and student with granularity $g$ of sample $x_j$. In this paper, we adopt three granularities $G = \{16,24,32\}$ as a default setting (See Table. \ref{table_region}). To further model the pixel-level information, we use MSE to maximize the output similarity predicted by the student and teacher models, and thus the loss is formulated as:
 		  \begin{equation}
 		  	\mathcal{L}_{pix} =\frac{1}{|\mathcal{D}_u|} \sum_{x_j \in \mathcal{D}_u} \left \| \mathcal{P}_j^t -\mathcal{P}_j^s \right \|_2^2
 		  \end{equation}
%\subsection{Semi-supervised Training Loss.} 
		 The overall training loss on labeled and unlabeled data is defined as:
		 \begin{equation}
		 	\mathcal{L}= \mathcal{L}_{sup}+ \lambda (\mathcal{L}_{fdc} + \mathcal{L}_{mrsc} + \mathcal{L}_{pix}),
		 \end{equation}
		 where  $\lambda$ is a time-dependent Gaussian warming up function \cite{yu2019uncertainty}.

\section{Experiments}
\subsection{Dataset and Evaluation Metrics.}
		\noindent \textbf{Datasets}. {Kvasir-SEG dataset} \cite{jha2020kvasir}  is designed for gastrointestinal polyp segmentation and includes 1000 polyp images along with their corresponding labels. We randomly choose 80\% of them as training set, and the remaining as testing set. {Skin lesion dataset} (ISIC 2016) \cite{gutman2016skin} contains dermoscopy images, which are used for automated diagnosis of skin cancer. It includes 900 training and 379 testing images.
		
		\noindent\textbf{Evaluation Metrics:} We evaluate performance using the following metrics: pixel-wise Accuracy (Acc), Mean Absolute Error (MAE), Dice Similarity Coefficient (Dice), and Intersection-over-Union (IoU).

\subsection{Implementation Details.}
 		The pre-trained backbone is SegFormer-B4 \cite{xie2021segformer}, and the adapter in  backbone is AdaptFormer \cite{chen2022adaptformer}. The AdamW optimizer with weight decay 0.0001 and momentum 0.9 is adopted for training 90 epochs, and the  learning rates is set to 0.0005. The images and labels are resized to 512$\times$512, with a batch size of 10. All experiments were conducted on an NVIDIA GeForce RTX 3090 GPU using the Pytorch library. 

\subsection{Comparisons with SOTA Methods.}
		We compare our method against recent SOTA methods on two datasets.  To make a fair comparison, we reimplemented these SOTA methods and obtained the results with the same backbone. Considering the reproducibility, we report the average results of 3 runs.
		The quantitative results are shown in Table \ref{table_sota}, when we only adopt the labeled data to train the supervised network, we achieve inferior performance than the SSL method, which indicates that the problem of insufficient annotation can be alleviated by utilizing the unlabeled data. Our FRCNet achieves 90.77\% Dice and 84.70\% IoU on skin dataset with 10\% labeled data, surpassing the second best method BCP \cite{bai2023bidirectional} by 1.75\% and 1.77\% in Dice and IoU scores.  Our method still outperforms the second best method ST++ \cite{yang2022st++} by 2.54\% and 2.20\% in Dice and IoU scores on polyp lesion segmentation task.  Table \ref{table_sota} shows all the previous SSL methods exhibit sub-optimal performance. In contrast, our method consider both region-level and frequency domain information, and outperforms all these methods.
		
\begin{figure}[h!]
      	\centering
      	\includegraphics[width=12cm]{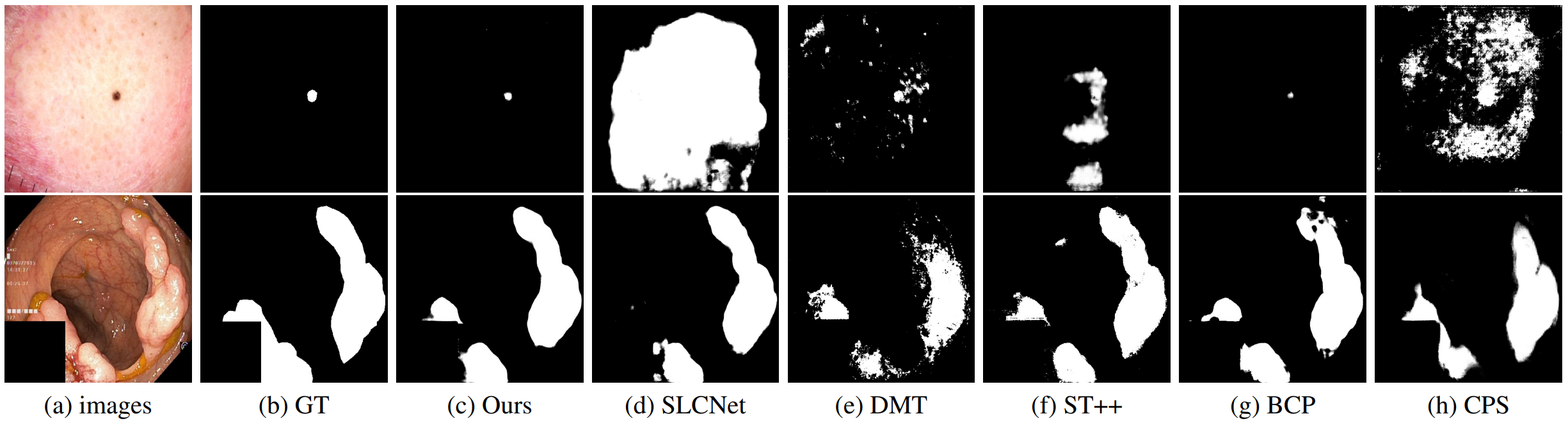}
      	\caption{The segmentation results of skin (first row) and polyp segmentation (second row) tasks using 10\% labeled data. }
      	\label{fig_visual}
      \end{figure}

		With 20\% labeled data, our method also achieved the best results, showing the effectiveness of the proposed consistency regularization strategies. Our approach can make use of large amounts of unlabeled data with less labeled data. When using only 20\% of the labeled data, the results are already very close to the supervised results.  For qualitative analysis, we show the visual comparisons of challenging cases in Fig. \ref{fig_visual}. It can be seen that the predictions of other methods fail to capture the details of small lesions (the first row) or can not be able to completely segment the edges of lesions (the second row), and our method has better scalability to lesions with different shapes and scales. 
		
	   \begin{table} [ht]
	   	\caption{Comparison of the proposed FRCNet and other SOTA SSL methods. Boldface and underline represent the best and second best results of each setting, respectively. }
	   	\scriptsize
	   	\centering
	   	\label{table_sota}
	   	\begin{tabular}{l|l|llll|lllll}
	   		\hline
	   		\multirow{2}{*}{Ratio} &\multirow{2}{*}{Method}& \multicolumn{4}{c|}{Kvasir-SEG}& \multicolumn{4}{c}{ISIC 2016}\\
	   		& &MAE$\downarrow$&Acc$\uparrow$& Dice$\uparrow$& IoU$\uparrow$   &MAE$\downarrow$&Acc$\uparrow$&Dice$\uparrow$& IoU$\uparrow$\\
	   		\hline
	   		100\%&SegFormer-B4 & \textbf{3.01 }  &  \textbf{97.48}  &  \textbf{91.79}  &  \textbf{86.86} & \textbf{3.37}      & \textbf{95.93}   & \textbf{92.88}   & \textbf{87.31}\\
	   		\hline
	   		\multirow{16}{*}{10\%}&SegFormer-B4      &  5.39   &  94.52  &  78.83  &  71.07 & 6.27      & 93.51   & 87.15   & 79.54\\
	   		\cline	{2-10}
	   		&MT \cite{tarvainen2017mean}                 &  4.82  &  94.97  &  83.45  &  76.32 &  5.96  &  94.03  &  88.01  &  81.42\\
	   		&SASSNet \cite{li2020shape}               &  4.91  &  94.78  &  82.78  &  75.21 &  5.84  &  93.78  &  88.42  &  81.46\\
	   		&CCT \cite{ouali2020semi}                   &  4.39  &  94.88  &  83.74  &  76.45 &  5.79  &  93.78  &  88.22  &  81.34\\
	   		&URPC  \cite{luo2021efficient}         &  4.89  &  94.69  &  82.43  &  75.49 &  5.74  &  93.74  &  88.12  &  81.21\\
	   		&DTC \cite{luo2021semi}                      &  4.98  &  94.60  &  80.75  &  73.84 &  5.82  &  93.62  &  87.94  &  81.23\\
	   		&CPS \cite{chen2021semi}                    &  4.67    &  94.91  &  82.81  &  75.34 &  5.23    &  94.28  &  88.89  &  82.12\\
	   		&SLCNet \cite{liu2022semi}                &  5.01  &  94.79  &  83.43  &  75.64 &  5.21  &  94.82  &  88.23  &  82.09\\
	   		&DMT \cite{feng2022dmt}                       &  4.34  &  94.89  &  84.98  &  77.23 &  4.94  &  94.93  &  88.03  &  82.34\\
	   		&ST++ \cite{yang2022st++}                    &  \underline{4.04}    &  \underline{95.31}  &  \underline{86.23}  &  \underline{80.41} &  5.03   &  94.98  &  88.17  &  82.46\\
	   		&BCP\cite{bai2023bidirectional}            &  4.21    &  95.09  &  85.79  &  78.43 & \underline{4.65}   & \underline{95.09}  &  \underline{89.02}  &  \underline{82.93}\\
	   		&FRCNet (Ours)  &  \textbf{3.51}   &  \textbf{96.49}  &  \textbf{88.77}  & \textbf{82.61}& \textbf{4.59}      & \textbf{95.41}   & \textbf{90.77}   & \textbf{84.70}\\
	   		\hline
	   		
	   		\multirow{16}{*}{20\%}
	   		&SegFormer-B4     & 4.86   &  95.27  &  82.16  &  74.69 & 5.76      & 94.32   & 88.65   & 81.58\\
	   		\cline	{2-10}
	   		&MT \cite{tarvainen2017mean}               &  4.67    &  95.47  &  84.45  &  77.43  &  5.05  &  95.43  &  89.83  &  83.15\\
	   		&SASSNet \cite{li2020shape}             &  4.45    &  95.56  &  83.97  &  77.21  &  5.03  &  95.19  &  89.94  &  83.67\\
	   		&CCT \cite{ouali2020semi}                   &  4.01    &  95.78  &  84.89  &  77.94  &  5.09  &  95.42  &  89.59  &  82.99\\
	   		&URPC  \cite{luo2021efficient}           &  4.35    &  95.52  &  85.12  &  78.21  &  5.03  &  95.21  &  89.91  &  82.81\\
	   		&DTC \cite{luo2021semi}                  &  4.31    &  95.65  &  84.22  &  78.01  &  5.11  &  95.39  &  89.79  &  83.32\\
	   		&CPS \cite{chen2021semi}               &  4.23     &  95.88  &  85.83  &  78.86  &  4.90  &  95.11  &  89.03  &  84.67\\
	   		&SLCNet \cite{liu2022semi}              &  4.81     &  95.74  &  85.42  &  78.35  &  4.86  &  95.34  &  89.34  &  84.54\\
	   		&DMT \cite{feng2022dmt}                 &  4.03     &  95.83  &  86.47  &  79.32  &  4.45  &  95.56  &  89.21  &  84.32\\
	   		&ST++ \cite{yang2022st++}                  &  \underline{3.47}     &  \underline{96.36 } &  \underline{87.94}  &  \underline{82.17}  &  \underline{4.12}  &  \underline{95.59}  &  \underline{90.21}  &  \underline{85.23}\\
	   		&BCP\cite{bai2023bidirectional}            &  {3.54}     &  {96.23}  & {87.43}  &  {80.45}  &  4.19  &  95.32  &  89.98  &  85.01\\
	   		&FRCNet (Ours)  &  \textbf{2.78}   &  \textbf{97.22}  &  \textbf{90.62}  & \textbf{85.01}& \textbf{3.70}      & \textbf{96.30}   & \textbf{92.44}   & \textbf{86.73}\\
	   		\hline
	   	\end{tabular}
	   \end{table}
      \vspace{-9mm}
      
\subsection{Ablation Study.}
	    
\begin{figure}[h!]
			\centerline{\includegraphics[width=4.8in]{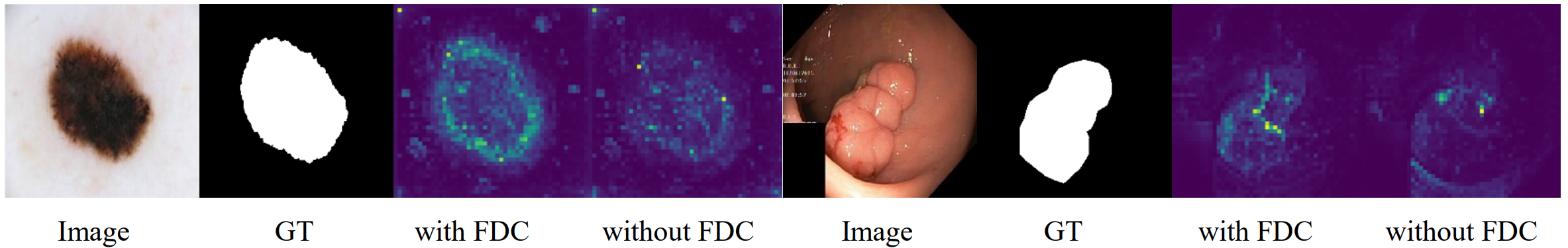}} 
			\caption{The feature response with and without frequency domain consistency regularization. We can see that after frequency domain feature learning, the  feature response of lesion region can be highlighted, and thus it can provide a more accurate decision boundary for the final segmentation.}
			\label{fig_Frequency}
			%\vspace{-3mm}
		\end{figure}

		\noindent \textbf{Analysis of FDC and MRSC.}  To study the importance of high frequency and low frequency for medical image segmentation, we integrated the two components with baseline, respectively. It can be seen from Table \ref{ablation} that low frequency component plays a more important role, because many lesions are similar to the color and texture features of surrounding normal regions, and they belong to the low frequency signals.  When high and low frequencies are integrated, the results can be further improved, indicating that the two frequency components are complementary. We visualized the features after considering frequency domain in  Fig. \ref{fig_Frequency}, and the results show that FDC can capture lesion information well by learning in frequency domain with FEM. 
		
		For MRSC, the performance is improved by more than 1.60\% in Dice and IoU scores and no trainable parameters are introduced. The best performance is achieved when FDC and MRSC are aggregated, outperforming the baseline by more than 5\% in Dice and IoU scores.  
		These above results fully demonstrate the effectiveness of our proposed method for semi-supervised medical image segmentation.  Finally,  {please note that our proposed FDC and MRSC are only used for training}. During the inference time, they can be removed and the computational complexity of inference is further reduced. 
		
		\noindent \textbf{Analysis of granularities in MRSC.} We study four granularity combinations  in Table \ref{table_region}, and they are $\{16\},\{16,24\}, \{16,24,32\}$ and $\{16,24,32,48\}$, respectively. The results show that MRSC with three granularities \{16,24,32\} can obtain the best performance. The results demonstrate that considering multiple region-level consistency can make the structural information and relationships among different image regions be effectively used, improving the feature representation ability of the model. 
		
		\noindent \textbf{Analysis of region projection in MRSC.} We can perform region projection in different ways, including linear interpolation, max pooling, average pooling, and convolution with different strides. We report the influence of region projection $\Psi_g$ in Table \ref{table_region}.  The results show that linear interpolation can achieve the best results, and max pooling is failed, the reason behind it is that max pooling only focuses on the maximum value, ignoring regions where the feature response is small but still important for lesion segmentation. Convolution can also achieve good results, but it will bring additional parameters. Based on the results, we choose linear interpolation as our default setting.
		
		    \begin{minipage}{0.4\textwidth}
			\scriptsize
			\renewcommand{\arraystretch}{1.2}
			\captionof {table} {Ablation studies of the proposed consistency regularizationon strategies on Kvasir-SEG dataset with 10\% labeled data.}
			\label{ablation}
			\begin{tabular}{l|llll}
				\hline
				Method& MAE$\downarrow$&       Acc$\uparrow$&    Dice$\uparrow$& IoU$\uparrow$\\
				\hline
				MT baseline                                     & 4.03 &94.97 &83.45 &76.32\\
				+ $\mathcal{L}^{low}_{fdc}$  			        & 3.98 &95.45 &86.01 &79.62\\
				+ $\mathcal{L}^{high}_{fdc}$  			        & 4.10 &95.21 &85.89 &78.91\\
				+ $\mathcal{L}_{fdc}$  					        & 3.98 &95.98 &86.84 &80.93\\
				+ $\mathcal{L}_{mrsc}$         			        & 3.95 &95.84 &86.17 &79.74\\
				+  $\mathcal{L}_{fdc}$ + $\mathcal{L}_{mrsc}$   &\textbf{3.51} &\textbf{96.49}&\textbf{88.77}&\textbf{82.61}\\
				\hline
			\end{tabular}
		    \end{minipage}
	        \hspace{0.8cm}
		    \begin{minipage}{0.5\textwidth}
			\scriptsize
			\captionof {table} {Analysis of different region granularities and projection operations in MRSC on Kvasir-SEG dataset with 10\% labeled data.}
			\label{table_region}
			\begin{tabular}{l|llllll}
				\hline
				Granularities& MAE$\downarrow$&       Acc$\uparrow$&    Dice$\uparrow$& IoU$\uparrow$\\
				\hline
				$\{16\}$                          & 3.89 &95.72 &87.69 &81.56\\
				$\{16,24\}$                       & 3.67 &96.13 &88.04 &82.02\\  
				$\{16,24,32\}$  		          & 3.51 &\textbf{96.49}&\textbf{88.77}&\textbf{82.61}\\
				$\{16,24,32,48\}$  		          & \textbf{3.34} &96.33 &88.23 &82.21\\ 
				\hline
				\hline
				Projection $\Psi_g$  & MAE$\downarrow$&       Acc$\uparrow$&    Dice$\uparrow$& IoU$\uparrow$\\
				\hline
				Linear  &    \textbf{3.51} &\textbf{96.49}&\textbf{88.77}&\textbf{82.61}\\
				AvgPool        &    3.69 &96.11 &88.01&82.01\\
				MaxPool           &    13.50 &86.50 &37.81 &25.45\\  
				Conv            &    4.01 &96.02 &87.43 &81.63\\
				\hline
			\end{tabular}
		\end{minipage}

\section{Conclusion}
	In this work, we propose a semi-supervised medical image segmentation framework FRCNet and its core components are the two  consistency regularization strategies FDC and MRSC, which are plug-and-play. FRCNet can improve the model’s capability for modeling consistency in both region-level and frequency domain. The results on two datasets reveal that it is critical to consider the consistency of the two domains, which allows the model to be trained efficiently with low labeled data regime and reduces the annotation cost by almost 80\%. These comprehensive experimental results verify the effectiveness of our proposed model.

\subsubsection{Acknowledgements} This work is partially supported by the National Natural Science Foundation (62272248), the Huazhu Fu’s Agency for Science, Technology and Research (A*STAR) Career Development Fund (C222812010) and Central Research Fund (CRF) and the Natural Science Foundation of Tianjin (23JCQNJC00010).

 \bibliographystyle{splncs04}
 \bibliography{mybib}

\begin{thebibliography}{10}
\providecommand{\url}[1]{\texttt{#1}}
\providecommand{\urlprefix}{URL }
\providecommand{\doi}[1]{https://doi.org/#1}

\bibitem{ahmed1974discrete}
Ahmed, N., Natarajan, T., Rao, K.R.: Discrete cosine transform. IEEE
  transactions on Computers  \textbf{100}(1),  90--93 (1974)

\bibitem{bai2023bidirectional}
Bai, Y., Chen, D., Li, Q., Shen, W., Wang, Y.: Bidirectional copy-paste for
  semi-supervised medical image segmentation. In: Proceedings of the IEEE/CVF
  Conference on Computer Vision and Pattern Recognition. pp. 11514--11524
  (2023)

\bibitem{chen2022adaptformer}
Chen, S., Ge, C., Tong, Z., Wang, J., Song, Y., Wang, J., Luo, P.: Adaptformer:
  Adapting vision transformers for scalable visual recognition. arXiv preprint
  arXiv:2205.13535  (2022)

\bibitem{chen2021semi}
Chen, X., Yuan, Y., Zeng, G., Wang, J.: Semi-supervised semantic segmentation
  with cross pseudo supervision. In: Proceedings of the IEEE/CVF Conference on
  Computer Vision and Pattern Recognition. pp. 2613--2622 (2021)

\bibitem{feng2022dmt}
Feng, Z., Zhou, Q., Gu, Q., Tan, X., Cheng, G., Lu, X., Shi, J., Ma, L.: Dmt:
  Dynamic mutual training for semi-supervised learning. Pattern Recognition
  \textbf{130},  108777 (2022)

\bibitem{Fu2018DiscAwareEN}
Fu, H., Cheng, J., Xu, Y., Zhang, C., Wong, D.W.K., Liu, J., Cao, X.:
  Disc-aware ensemble network for glaucoma screening from fundus image. IEEE
  Transactions on Medical Imaging  \textbf{37},  2493--2501 (2018)

\bibitem{gutman2016skin}
Gutman, D., Codella, N.C., Celebi, E., Helba, B., Marchetti, M., Mishra, N.,
  Halpern, A.: Skin lesion analysis toward melanoma detection: A challenge at
  the international symposium on biomedical imaging (isbi) 2016, hosted by the
  international skin imaging collaboration (isic). arXiv preprint
  arXiv:1605.01397  (2016)

\bibitem{jha2020kvasir}
Jha, D., Smedsrud, P.H., Riegler, M.A., Halvorsen, P., Lange, T.d., Johansen,
  D., Johansen, H.D.: Kvasir-seg: A segmented polyp dataset. In: International
  Conference on Multimedia Modeling. pp. 451--462. Springer (2020)

\bibitem{Kirkpatrick2016OvercomingCF}
Kirkpatrick, J., Pascanu, R., Rabinowitz, N.C., Veness, J., Desjardins, G.,
  Rusu, A.A., Milan, K., Quan, J., Ramalho, T., Grabska-Barwinska, A.,
  Hassabis, D., Clopath, C., Kumaran, D., Hadsell, R.: Overcoming catastrophic
  forgetting in neural networks. Proceedings of the National Academy of
  Sciences  \textbf{114},  3521 -- 3526 (2016)

\bibitem{li2020shape}
Li, S., Zhang, C., He, X.: Shape-aware semi-supervised 3d semantic segmentation
  for medical images. In: International Conference on Medical Image Computing
  and Computer-Assisted Intervention. pp. 552--561. Springer (2020)

\bibitem{li2018h}
Li, X., Chen, H., Qi, X., Dou, Q., Fu, C.W., Heng, P.A.: H-denseunet: hybrid
  densely connected unet for liver and tumor segmentation from ct volumes. IEEE
  transactions on medical imaging  \textbf{37}(12),  2663--2674 (2018)

\bibitem{li2020transformation}
Li, X., Yu, L., Chen, H., Fu, C.W., Xing, L., Heng, P.A.:
  Transformation-consistent self-ensembling model for semisupervised medical
  image segmentation. IEEE Transactions on Neural Networks and Learning Systems
   \textbf{32}(2),  523--534 (2020)

\bibitem{liu2022semi}
Liu, J., Desrosiers, C., Zhou, Y.: Semi-supervised medical image segmentation
  using cross-model pseudo-supervision with shape awareness and local context
  constraints. In: International Conference on Medical Image Computing and
  Computer-Assisted Intervention. pp. 140--150. Springer (2022)

\bibitem{luo2021semi}
Luo, X., Chen, J., Song, T., Wang, G.: Semi-supervised medical image
  segmentation through dual-task consistency. In: Proceedings of the AAAI
  Conference on Artificial Intelligence. vol.~35, pp. 8801--8809 (2021)

\bibitem{luo2021efficient}
Luo, X., Liao, W., Chen, J., Song, T., Chen, Y., Zhang, S., Chen, N., Wang, G.,
  Zhang, S.: Efficient semi-supervised gross target volume of nasopharyngeal
  carcinoma segmentation via uncertainty rectified pyramid consistency. In:
  International Conference on Medical Image Computing and Computer-Assisted
  Intervention. pp. 318--329. Springer (2021)

\bibitem{ouali2020semi}
Ouali, Y., Hudelot, C., Tami, M.: Semi-supervised semantic segmentation with
  cross-consistency training. In: Proceedings of the IEEE/CVF Conference on
  Computer Vision and Pattern Recognition. pp. 12674--12684 (2020)

\bibitem{ronneberger2015u}
Ronneberger, O., Fischer, P., Brox, T.: U-net: Convolutional networks for
  biomedical image segmentation. In: International Conference on Medical image
  computing and computer-assisted intervention. pp. 234--241. Springer (2015)

\bibitem{sohn2020fixmatch}
Sohn, K., Berthelot, D., Carlini, N., Zhang, Z., Zhang, H., Raffel, C.A.,
  Cubuk, E.D., Kurakin, A., Li, C.L.: Fixmatch: Simplifying semi-supervised
  learning with consistency and confidence. Advances in Neural Information
  Processing Systems  \textbf{33},  596--608 (2020)

\bibitem{tarvainen2017mean}
Tarvainen, A., Valpola, H.: Mean teachers are better role models:
  Weight-averaged consistency targets improve semi-supervised deep learning
  results. Advances in neural information processing systems  \textbf{30}
  (2017)

\bibitem{vaswani2017attention}
Vaswani, A., Shazeer, N., Parmar, N., Uszkoreit, J., Jones, L., Gomez, A.N.,
  Kaiser, {\L}., Polosukhin, I.: Attention is all you need. Advances in neural
  information processing systems  \textbf{30} (2017)

\bibitem{Wu2021AutomatedSL}
Wu, H., Pan, J., Li, Z., Wen, Z., Qin, J.: Automated skin lesion segmentation
  via an adaptive dual attention module. IEEE Transactions on Medical Imaging
  \textbf{40},  357--370 (2021)

\bibitem{wu2021semi}
Wu, Y., Xu, M., Ge, Z., Cai, J., Zhang, L.: Semi-supervised left atrium
  segmentation with mutual consistency training. In: International Conference
  on Medical Image Computing and Computer-Assisted Intervention. pp. 297--306.
  Springer (2021)

\bibitem{xie2021segformer}
Xie, E., Wang, W., Yu, Z., Anandkumar, A., Alvarez, J.M., Luo, P.: Segformer:
  Simple and efficient design for semantic segmentation with transformers.
  Advances in Neural Information Processing Systems  \textbf{34},  12077--12090
  (2021)

\bibitem{yang2022st++}
Yang, L., Zhuo, W., Qi, L., Shi, Y., Gao, Y.: St++: Make self-training work
  better for semi-supervised semantic segmentation. In: Proceedings of the
  IEEE/CVF Conference on Computer Vision and Pattern Recognition. pp.
  4268--4277 (2022)

\bibitem{yu2019uncertainty}
Yu, L., Wang, S., Li, X., Fu, C.W., Heng, P.A.: Uncertainty-aware
  self-ensembling model for semi-supervised 3d left atrium segmentation. In:
  International Conference on Medical Image Computing and Computer-Assisted
  Intervention. pp. 605--613. Springer (2019)

\bibitem{zhong2022detecting}
Zhong, Y., Li, B., Tang, L., Kuang, S., Wu, S., Ding, S.: Detecting camouflaged
  object in frequency domain. In: Proceedings of the IEEE/CVF Conference on
  Computer Vision and Pattern Recognition. pp. 4504--4513 (2022)

\bibitem{zhou2018unet++}
Zhou, Z., Rahman~Siddiquee, M.M., Tajbakhsh, N., Liang, J.: Unet++: A nested
  u-net architecture for medical image segmentation. In: Deep learning in
  medical image analysis and multimodal learning for clinical decision support,
  pp. 3--11. Springer (2018)

\end{thebibliography}
\end{document}